\documentclass[11pt,a4paper]{article}
\usepackage[hyperref]{acl2021}
\usepackage{times}
\usepackage{latexsym}

\usepackage{microtype}
\usepackage{makecell}
\usepackage{arydshln}
\usepackage{xcolor}
\usepackage{tcolorbox}
\usepackage{multirow}
\usepackage{graphicx}
\usepackage{amsmath}
\usepackage{booktabs}

\aclfinalcopy % Uncomment this line for the final submission
\title{Affective Decoding for Empathetic Response Generation}

\author{Chengkun Zeng\textsuperscript{$\spadesuit$}, 
Guanyi Chen\textsuperscript{$\heartsuit$}, Chenghua Lin\textsuperscript{$\spadesuit$}\thanks{~~Corresponding author}~, Ruizhe Li\textsuperscript{$\spadesuit$}, Zhigang Chen\textsuperscript{$\clubsuit$}\\
\textsuperscript{$\spadesuit$}Department of Computer Science, University of Sheffield\\
\textsuperscript{$\heartsuit$}Department of Information and Computing Sciences, Utrecht University\\
\textsuperscript{$\clubsuit$}College of Information and Intelligent Engineering, Zhejiang Wanli University\\%, China\\
\texttt{chengkunzenggo@gmail.com,}
\texttt{\{c.lin, r.li\}@sheffield.ac.uk}\\ 
\texttt{g.chen@uu.nl, chenzhigang@zwu.edu.cn}}

\date{}

\begin{document}
\maketitle
\begin{abstract}
Understanding speaker's feelings and producing appropriate responses with emotion connection is a key communicative skill for empathetic dialogue systems. In this paper, we propose a simple technique called Affective Decoding for empathetic response generation. Our method can effectively incorporate emotion signals during each decoding step, and can additionally be augmented with an auxiliary dual emotion encoder, which learns separate embeddings for the speaker and listener given the emotion base of the dialogue. Extensive empirical studies show that our models are perceived to be more empathetic by human evaluations, in comparison to several strong mainstream methods for empathetic responding.

\end{abstract}

\section{Introduction} \label{sec:intro}

Endowing a dialogue system with the ability of empathy responding has attracted a growing body of research~\cite{ma2020survey} and is believed to be crucial for many service-oriented applications, such as mental health interventions~\cite{hoermann2017application}, assisting medical diagnosis~\cite{xu2019end}, and restaurant/hotel booking services~\cite{ghazvininejad2017knowledge,liu-etal-2018-dialogue,wang-etal-2021-fast}.
Being empathetic requires one to be able to understand the implied feeling of the conversation partner, or in other words, to place oneself in partner's position. Therefore, to produce proper responses, an Empathetic Dialogue System (EDS) needs to understand not only the situation of the speaker\footnote{We use the term ``speaker'' referring to the user of the empathetic dialogue system while ``listener'' inferring to the dialogue system itself.} and the causes~\cite{abd-yusof-etal-2017-analysing},  
but also the emotion state of speaker. 
%in light of the experience depicted by the speaker.

% Each dialogue session is initialised by speaker's description of his experience about breaking up with his girl friend, grounding on which the listener understands his emotion (i.e., lonely in this case) and responses empathetically by uttering \emph{Sorry to hear that!} and \emph{Sorry again!}.

\begin{figure}[t]
    \centering
    \includegraphics[width=0.8\columnwidth]{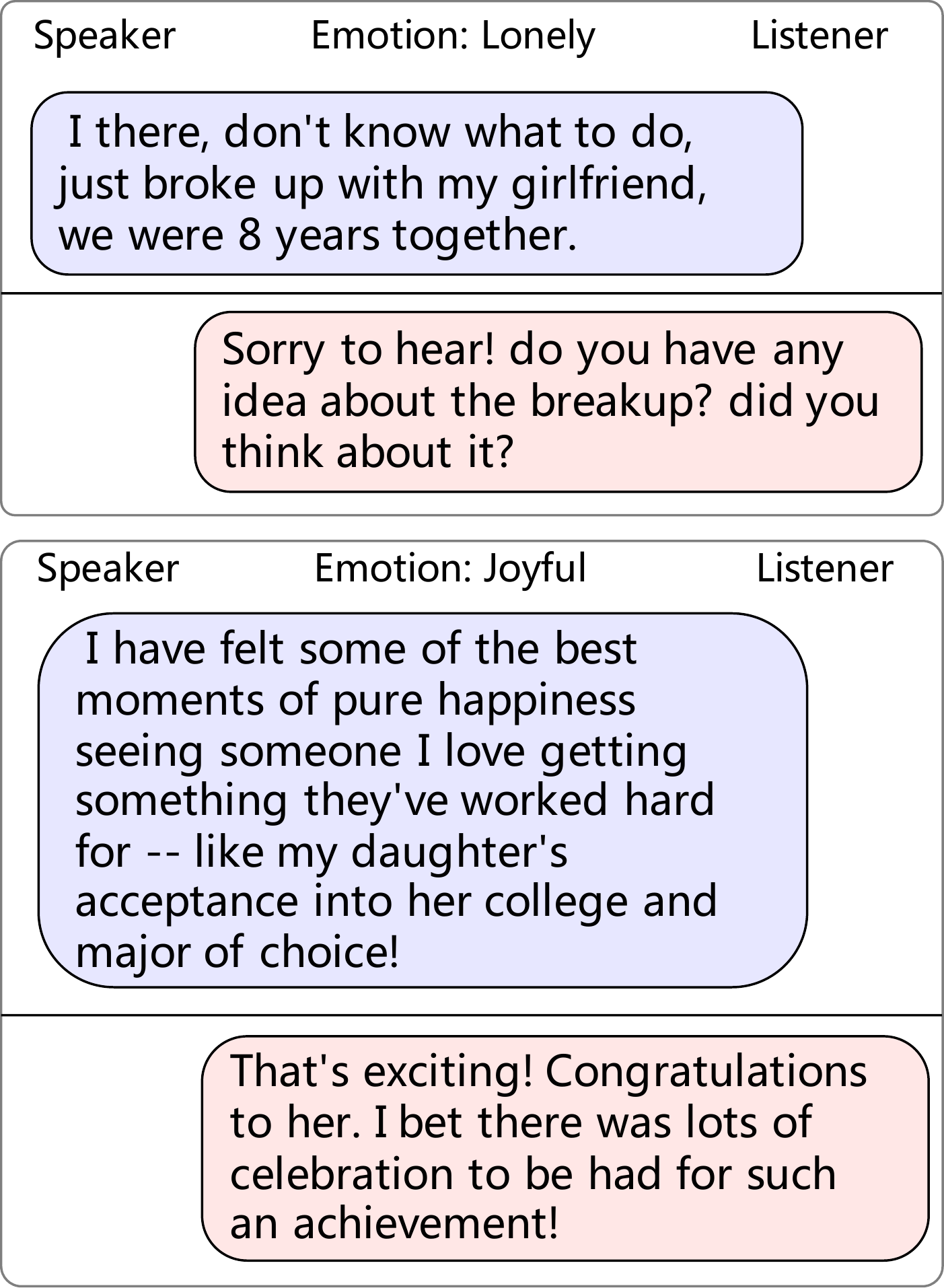}
    \caption{Two example dialogue sessions from the \textsc{EmpDial} dataset with the emotion \textit{lonely} and \textit{joyful}, repsectively.}
    \label{fig:example}
\end{figure}

\begin{figure*}[t]
    \centering
    \includegraphics[width=1.3\columnwidth]{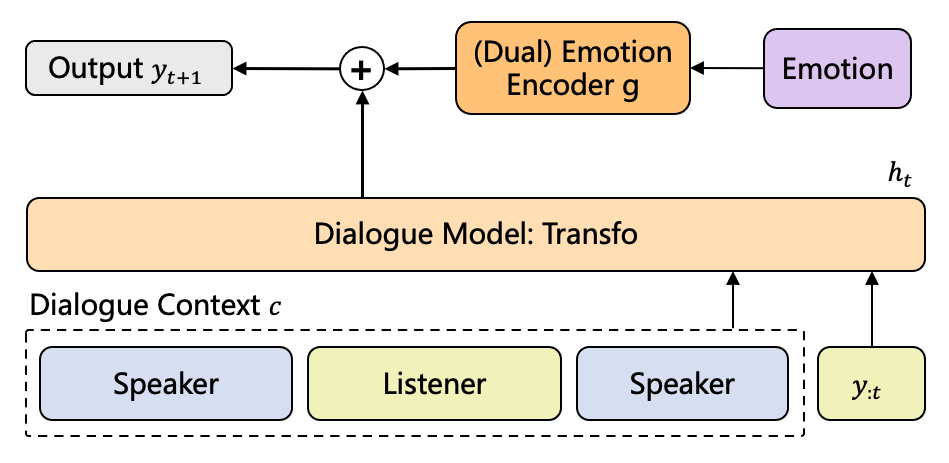}
    \caption{The overall architecture of our empathetic dialogue system.}
    \label{fig:model}
\end{figure*}
In 2019, \citet{Rashkin2019d} formally introduced the task for dialogue systems to respond to conversations with emotions. They also constructed a benchmark corpus called \textsc{EmpatheticDialogues} (abbreviated as \textsc{EmpDial}), which consists of conversations with a wide range of emotion states for task evaluation. 
Figure~\ref{fig:example} shows an example session of a dialogue from \textsc{EmpDial}, where the situation reflects the emotion state of \textit{lonely} and \textit{joyful}.
Several approaches have been proposed for modelling emotions, which is a key challenge for building an EDS. 
These approaches follow two main enterprises: 
one is multi-task learning~\cite{Rashkin2019d,lin2020caire}, which trains models for both dialogue generation and predicting the emotion of the dialogue; 
the other enforces the model to generate empathetic responses conditioning on the emotion state predicted from the dialogue context with a pre-trained emotion classifier~\cite{Rashkin2019d}.

Our work follows a similar vein to the second enterprise, where we propose a simple yet efficient technique coined \emph{Affective Decoding} (AD), which can effectively incorporate emotion signals into model training and generate more empathetic responses. 
Our method can work in two different modes. The first mode injects emotion embedding in each decoding step. This is different to~\citet{Rashkin2019d} who only applied prepending at the first time-step on the encoder. For the second mode, we introduce an additional auxiliary \emph{Dual Emotion Encoder}, which learns separate embeddings for the speaker and listener given the emotion base of the dialogue. 
In addition, we systematically evaluate and compare AD with the existing mainstream emotion modelling methods for empathetic responding, including both prepending emotion embeddings~\cite{Rashkin2019d} and multi-task learning~\cite{lin2020caire}.

Based on the \textsc{EmpDial} dataset, we conducted comprehensive empirical studies, which include automatic (e.g., BLEU, BOW) and two human evaluations. 
For human evaluation, we assess both model-level performance and finer-grained level aspects concerning empathy, relevance, and fluency of the generated responses. 
Empirical results show the effectiveness of our affective decoding and that our model with the auxiliary dual emotion encoder works the best. 
While~\citet{Rashkin2019d} reported that multi-task learning did not provide consistent improvements for the task, we actually found multi-task learning performs even worse than a pre-trained language model~\cite{Wolf2019g} fine-turned on the \textsc{EmpDial} dataset. 

To summarise, our contributions are 3-fold: (1) we introduce a simple yet efficient decoding method called affective decoding to the task of empathetic response generation; (2) we conduct a comprehensive comparison between various emotion modelling methods in empathetic dialogue modelling by means of automatic evaluation and 2 human evaluations; (3) empirical results show the effectiveness of our affective decoding method and that with the auxiliary dual emotion encoder, our model can further support the analysis of listeners and  speakers' behaviours  in  terms  of  how  they  utter  with  respected  to  the  same  emotion.

The rest of the paper is organised as follows. 
%\S2 introduces the related work. 
\S2 presents our model for empathetic response generation. We show the experimental setup and results in \S3. \S4 presents some case studies and finally \S5 concludes the paper.
The code is available at: \url{https://github.com/zenggo/affective-decoding-4-empathetic-dialog}.

%\textcolor{red}{To summarise, our contributions are x-fold: [TODO]}

\begin{figure}[tb]
    \centering
    \includegraphics[width=\columnwidth]{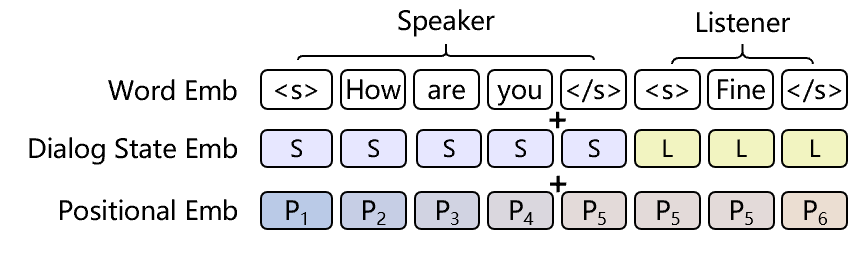}
    \caption{Illustration of the input format of our model.}
    \label{fig:input}
\end{figure}

\section{Methodology} \label{method}

In this section, we describe our Affective Decoding model, which consists of two key components, namely, a pre-trained response generator, \textit{Transfo}~\cite{Wolf2019g}, and the affective decoder for enhancing empathetic responding.

\subsection{Dialogue Modelling}

We use Transfo, which is built upon the Generative Pre-trained Transformer~\cite[GPT]{Radford2018b} pre-trained on the \textsc{BooksCorpus} dataset~\cite{Zhu2015}, and which gives good performance on building conversational agents~\cite{dinan2020second}. 
When fine-tuning on the \textsc{EmpDial} dataset, a response is generated given the dialogue context $c$, which contains single or multi-turn conversations. 
For each input token, it is represented as a summation of its word embedding, positional embedding and dialogue state embedding, as illustrated in Figure~\ref{fig:input}. 
We model two possible dialogue states, where state $S$ corresponds to the speaker and state $L$ to the listener. % \textcolor{orange}{[Chen: maybe explain a bit why modeling S and L?]}  % \RED{[CL: say something about why the fine tuning stage is important here.]}
%\BLUE{[CZ: During generation, the dialog state at time $t$ can be arbitrarily set to S or L so that the model can act as either speaker or listener]} 
%As an EDS, though our model trains both speaker and listener, but only the listener is evaluated.
 
\begin{table*}[t]
%\resizebox{\columnwidth}{!}{%
    \centering
    \begin{tabular}{lcccc}
    \toprule
        Dataset & Emotion Labels & Train & Validation & Test \\
    \midrule
    \textsc{EmpDial} & 32 & 19,533 & 2,770 & 2,547 \\
    \bottomrule
    \end{tabular}
 %   }
    \caption{The statistics of \textsc{EmpDial} dataset.}
    \label{tab:data_sta}
\end{table*}

\subsection{Affective Decoding}

One of the key challenges of building an effective EDS is recognising and understanding the emotion of the speaker. 
%where the important signal is the treatment of ... 
Inspired by the affective language model of \citet{Ghosh2017d}, we tackle the problem by proposing \textit{Affective Decoding} (\textbf{AD}), a simple strategy which injects emotion embeddings into each decoding step. Such a strategy allows our model to encode dialogue's emotion base effectively, and to distribute more probability mass towards the words in the utterance that are highly correlated with the dialogue emotion, leading to enhanced empathetic responding. 
%this process is to adjust the probabilities of specific words (which are highly correlated to the emotion) appearing in the utterance
%Note that in the \textsc{EmpDial} dataset, there is a gold standard emotional situation label for each dialogue. 

Concretely, at each time step $t$, we first encode the emotional label with a one-hot embedding, which is then mapped into a dense vector $g(\mathbf{e})$ by the emotion encoder $g(\cdot)$ (see Figure~\ref{fig:model} for details). 
$g(\mathbf{e})$ is then used for predicting the next word $y_{t+1}$ jointly with the dialogue context $c$ and the decoded outputs for all previous time steps $y_{:t}$. Formally, the probability of $P(y_{t+1})$ is given as
\begin{equation}
P(y_{t+1}| y_{:t}, c, e) = \mbox{softmax}(W h_t + Vg(\mathbf{e})),
\end{equation}
where $h_t$ is the representation of $c$ and $y_{:t}$ encoded by \textit{Transfo}; $W$ and $V$ are weights in the output layer. 
Similar to prior studies~\cite{Rashkin2019d,lin2020caire}, our  AD model maintains one emotion embedding for the whole dialogue session. 

\noindent\textbf{Dual Emotion encoder.}~~We observe that in the dialogues with emotional situations, speaker and listener tend to utter with distinctive styles. That is, the speaker normally describes his/her own experience with personal emotions, while the listener tries to respond in the way which can establish an emotional connection with the speaker based on speaker's emotional needs (e.g., encouraging and motivating). 
%establish emotional connection with the speaker 
%is heading to react in an empathetic way with respected to that  emotion. 
For example, in the dialogue with a emotional base of \textit{joyful} in Figure~\ref{fig:example}, the speaker used phrases like \emph{``happiness''} and \emph{``love''} while listeners used \emph{``exciting''} and \emph{``congratulation''}.
Based on this observation, we introduce a mechanism so called Dual Emotion (DE) encoder, which learns separate embeddings for the speaker and listener given the emotion base of the dialogue. 
%another mode of our AD by incorporating Dual Emotion (DE) encoder which maps the emotion to different embedding space for disassociating speakers and listeners. 
We coin our model augmented with the auxiliary DE component \textbf{AD+DE}, and its generation process becomes:
\begin{equation}\label{eq:softmax}
\begin{split} 
    &P(y_{t+1}| y_{:t}, c, e) = \\
    &\begin{cases}
    \text{softmax}(W h_t + V_{\scriptscriptstyle S} g_{\scriptscriptstyle S}(\mathbf{e})) & s_t = S\\
    \text{softmax}(W h_t + V_{\scriptscriptstyle L}g_{\scriptscriptstyle L}(\mathbf{e})) & s_t = L,
    \end{cases}
\end{split}
\end{equation}
where $s_t \in \{S, L\}$ is the dialogue state at the step $t$. With the dual embedding space, we hypothesise that the interpretability of our model's behaviour will also be enhanced, as it makes possible to identify the differences of the language use between speakers and listeners.

\section{Experiment}
\begin{table*}[t]
%\resizebox{\columnwidth}{!}{%
    \centering
    \begin{tabular}{lccccccc}
    \toprule
        Model & PPL & BLEU & BOW$_e$ & BOW$_a$ & BOW$_g$ & DIST-1 & DIST-2\\
    \midrule
    MoEL & 38.19 & \textbf{2.84} & 0.502 & \textbf{0.878} & \textbf{0.679} & 0.005 & 0.023 \\
    %Transfo (w/o pt) & 31.21 & 1.26 & 0.303 & 0.608 & 0.415 & 0.005 & 0.023\\
    %\hdashline
    Transfo               & 13.94 & 1.75 & 0.729 & 0.747 & 0.559 & 0.015 & 0.070\\
    Prepend               & \textbf{13.90} & 1.75 & 0.731 & 0.753 & 0.565 & \textbf{0.016} & \textbf{0.079}\\
    MTL                   & 14.95 & 1.49 & 0.733 & 0.757 & 0.566 & 0.015 & 0.067\\\midrule
    ADM             & 14.50 & 1.81 & 0.733 & 0.756 & 0.564 & 0.015 & 0.068\\
    AD                    & 14.04 & 1.69 & 0.734 & 0.756 & 0.570 & \textbf{0.016} & 0.072\\
    AD+DE                 & 14.03 & 1.71 & \textbf{0.736} & 0.757 & 0.571 & \textbf{0.016} & 0.074\\
    \bottomrule
    %     TA-Transfo &116.3M & 15.45 &\textbf{1.86}\\
    % \hline
    \end{tabular}%}
    \caption{Automatic evaluation results.} % Transfo (w/o pt) means Transfo without pre-training}
    \label{tab:auto_result}
\end{table*}

We evaluate our models on \textsc{EmpDial}, using the original split of \citet{Rashkin2019d} and their emotion classifier based on FastText~\citep{joulin2017bag}. Table~\ref{tab:data_sta} shows the statistics of the \textsc{EmpDial} dataset, which contains 32 emotion labels.
%which is splitted and used the same as in~\citet{Rashkin2019d}. We also follow \citet{Rashkin2019d} who applied an emotion classifier based on fastText. 
We compare our model against four competitive baselines in the experiment, including 
\begin{itemize}
    \item \textbf{MoEL}: a transformer model with multiple independent transformer decoders for generating different contextual responses~\citep{lin2019moel};
    \item \textbf{Transfo}: a pre-trained transformer model which is fine-tuned using multi-task learning in language modelling and next-utterance classification tasks~\cite{Wolf2019g};
    \item \textbf{PRE}: a Transfo model with an emotion embedding prepended to the dialogue context~\cite{Rashkin2019d};
    \item \textbf{MTL}: a Transfo model with multi-task learning, where the main task is dialogue response generation and the secondary task uses the encoded dialogue context for predicting the emotion for the whole session~\cite{lin2020caire}. 
\end{itemize}

In terms of our own models, apart from \textbf{AD} and \textbf{AD+DE}, we also further tested a model variant (\textbf{ADM}), which considers multi-task learning. We detail each of our model variants below.
\begin{itemize}
    \item \textbf{AD}: a simple model by injecting emotion embeddings into each decoding step;
    \item \textbf{AD+DE}: a variant of \textbf{AD} by introducing the Dual Emotion encoder to separately model embeddings for the speaker and listener;
    \item \textbf{ADM}: a variant of our model by combining \textbf{AD+DE} with multi-task learning, adopting a similar strategy to the \textbf{MTL} baseline. 
\end{itemize}
% Apart from the two variants of our models described in \S \ref{method} (i.e. AD and AD+DE),  we further test a variant by combining AD+DE with multi-task learning (\textit{ADM}), adopting a similar strategy to the \textit{MTL} baseline. 
%Besides, it is not fair to directly compare MTL with other methods as it does not need emotion label in the inference phrase. Therefore, we also try the model combining AD+DE with MTL, resulting \textbf{ADM}.

%As aforementioned in section~\ref{sec:intro}, we test Multi-Task Learning (\textbf{MTL}). To be specific, the dialogue context $c$ is encoded using \textit{Transfo} and, subsequently, is used for predicting emotion of the conversation $\hat{\mathbf{e}}$. The overall objective function is augmented as: $\mathcal{L} = \mathcal{L}_r + \mathcal{L}_e$, where $\mathcal{L}_r$ is the loss for response generation while $\mathcal{L}_e$ is the loss for the emotion classifier. It is not fair to directly compare MTL with other methods as it does not need emotion label in the inference phrase. Therefore, we also try combine MTL and AD+DE (i.e., \textbf{ADM}).

\subsection{Automatic Evaluation}

For automatic evaluation, we evaluate the models in three aspects, i.e., \textit{fluency}, \textit{adequacy}, and \textit{diversity}. Particularly, fluency is measured by perplexity, adequacy by BLEU and BOW embedding metrics, and diversity by DIST. We describe each of the metrics in detail below.

\begin{itemize}
    \item  \textbf{Perplexity (PPL)}: measures how well a language model is (lower the better);
    \item 
    \textbf{BLEU}~\cite{papineni2002bleu}: $n$-gram overlap between the system output and the reference;\\
    \textbf{BOW embedding}~\cite{liu2016not}: the cosine similarity between the bag-of-words embeddings of the output and the reference. Specifically, there are three matching strategies:
        \begin{itemize}
            \item Greedy (\textbf{BOW$_g$}): the average cosine similarities between word embeddings of the two utterances which are greedily matched~\citep{rus2012optimal};
            \item Average (\textbf{BOW$_a$}):  the cosine similarity between the averaged word embeddings in the two utterances~\citep{mitchell2008vector};
            \item Extreme (\textbf{BOW$_e$}):  the cosine similarity between the largest extreme values in the word embeddings of the two utterances~\citep{pennington2014glove};
        \end{itemize}
   
    \item \textbf{DIST}~\cite{li2016diversity}: measures the corpus level diversity of the outputs by calculating the ratio of unique $n$-grams ($n=1,2$) over all $n$-grams in the outputs.
    
\end{itemize}

%we use two word overlap measures,  
% by \textbf{BLEU}~\cite{papineni2002bleu} (i.e., $n$-gram overlap between the system output and the reference) and \textbf{BOW embedding}~\cite{liu2016not} (i.e., the cosine similarity between the bag-of-words embeddings of the output and the reference with three matching strategies, including greedy (\textbf{BOW$_g$}), average (\textbf{BOW$_a$}) and extreme (\textbf{BOW$_e$})); 3) \emph{Diversity}: by \textbf{DIST}~\cite{li2016diversity} which measures the corpus level diversity of the outputs by calculating the ratio of unique $n$-grams ($n=1,2$) over all $n$-grams in the outputs.

\subsubsection{Experimental Results}

Table~\ref{tab:auto_result} shows the automatic evaluation results for the tested models. 
Overall, the results do not seem to provide strong evidence in terms of which models perform best. 
Among the baselines, MoEL achieves the highest BLEU, BOW$_a$ and BOW$_g$, while it has the worst PPL and diversity (i.e., DIST-1 and DIST-2). Prepend in contrast, performs the best in terms of PPL and diversity, and gives similar performance in the BOW metrics when compared to other baselines (except MoEL) and our models.

Our AD+DE model gives similar performance to Prepend, i.e., it achieves fairly balanced performance across all types of metrics and gives the highest scores in BOW$_e$ and DIST-1. AD+DE also appears to slightly outperform AD, but the difference is somewhat minimal.
%outperforms other models with a small margin (i.e., it achieves the highest score in two metrics as well as the second highest scores in DIST-2). Also the difference between AD+DE and AD is somewhat minimal. 
%Whether its victories exist and have statistical significance need to be validated by human judges.
%the automatic evaluation results provides no clear evidence for a conclusion indicating one variant is more favourable than another. Nonetheless, by synthesising all evaluation metrics, we see the AD+DE or the ADM might outperform other models by narrow margins.
%\RED{So far, it is hard to understand how each of these metrics would reflect on the overall performance of a EDS}, thus we could not access the how well it works comparing to othere models. 
%\textbf{Second}, by comparing the transfo without pre-training and the rest, we found that pre-training really help with generating good responses in the EDS.
In addition, it is surprising to see no significant difference between Transfo and other models for all metrics, where the latter explicit account for the emotional signals of the dialogue. 
We also see that MTL has a lower BLEU score even than Transfo. 
Conversely, comparing AD+DE and ADM, multi-task learning helps to yield better BLEU but yields worse performance on other metrics. 
In summary, we are not able to establish a clear winner based on automatic metrics, although Prepend seems to slightly outperform other baseline models overall. 
%we are not able to obtain sensible model comparisons based on automatic metrics. 

\begin{table*}[t]
% \resizebox{\textwidth}{!}{%
    \centering \small
    \begin{tabular}{lccccccc}
\toprule
        Model  & MoEL & Transfo  & PRE  & MTL  & ADM  & AD & AD+DE \\\midrule
        % MoEL & - & 12~ & 11~ & 26~ & 18~ & 11~ & 12 \\
        MoEL & - & 12~ & 11~ & 26~ & 18~ & 11~ & 12\\
        Transfo & 48$^\dag$ & - & 23~ & 28~ & 29 & 13~ & 17 \\
        PRE & 52$^\dag$ & 34$^\dag$ & - & 49$^\dag$ & 34 & 19~ & 22 \\
        MTL & 48$^\dag$ & 25~ & 19~ & - & 27~ & 16~ & 15 \\
        ADM & 56$^\dag$ & 45$^\dag$ & 26~ & 46$^\dag$ & - & 22~ & 21 \\
        AD & 55$^\dag$ & 39$^\dag$ & 45$^\dag$ & 44$^\dag$ & 40$^\dag$ & -  & 8 \\
        AD+DE & 57$^\dag$ & 42$^\dag$ & 35$^\dag$ & 45$^\dag$ & 43$^\dag$ & 12~ & - \\ 
    \bottomrule
    \end{tabular}
    % }
    \caption{Results of the pairwise preference test: the number indicates  the percentage (\%) of responses generated by system A (row) is favoured by raters comparing to B (column). $^\dag$ means the preference is significant with $p<.05$ (two-proportion z-test).}
    \label{tab:pt}
\end{table*}

\begin{table*}[tb!]
    \centering \small
    \begin{tabular}{clcccccccccccccccc}
    \toprule
        & & \multicolumn{7}{c}{Single-turn} & & & \multicolumn{7}{c}{Multi-turn} \\
        \cmidrule(lr){3-9}\cmidrule(lr){12-18}
        & Model & 1 & 2 & 3 & 4 & 5 & 6 & 7 & & & 1 & 2 & 3 & 4 & 5 & 6 & 7\\\midrule
        1 & MoEL & - & 15 & 12 & 29 & 19 & 14 & 16 & & & - & 9 & 10 & 23 & 17 & 8 & 8\\
        2 & Transfo & 46 & - & 14 & 28 & 36 & 14 & 22 & & & 50 & - & 32 & 28 & 22 & 12 & 12\\
        3 & Prepend & 47 & 42 & - & 46 & 44 & 24 & 20 & & & 57 & 26 & - & 52 & 24 & 14 & 24\\
        4 & MTL & 42 & 32 & 16 & - & 26 & 16 & 16 & & & 54 & 18 & 22 & - & 28 & 16 & 14\\
        5 & MTL+AD+DE & 52 & 44 & 22 & 44 & - & 28 & 28 & & & 60 & 46 & 30 & 48 & - & 16 & 14\\
        6 & AD & 49 & 52 & 40 & 48 & 46 & - & 16 & & & 61 & 26 & 50 & 40 & 34 & - & 8\\
        7 & AD+DE & 51 & 52 & 32 & 42 & 48 & 10 & - & & & 63 & 32 & 38 & 48 & 38 & 6 & -\\
    \bottomrule
    \end{tabular}
    \caption{Results of the pairwise preference test for single-turn EDS (left) and multi-turn EDS (right): the number indicates the percentage (\%) of responses generated by system A (row) is favoured by raters comparing to B (column).}
    \label{tab:pt_full}
\end{table*}

\subsection{Human Evaluation}

%These observations somewhat reflect that automatic metrics do not provide good measurements of generation quality. 

To assess the performance of the tested models more robustly and comprehensively, 
%and to understand the performance from different angles, 
we conducted two forms of human evaluation: \emph{ranking} for evaluating the overall performance of each system~\cite{duh2008ranking}, and \emph{multi-item rating}~\cite{diamantopoulos2012guidelines} for evaluating the system performance against more fine-grained aspects (e.g. whether the response is relevant or not).

%\textcolor{red}{[CL: can we add more details about the human evaluation settings please, as suggested by the reviewers.]}
%rate each system output with different criteria for more break-down assessment, especially whether the response is empathy or not).
%(i.e., rank each system output for better evaluating the overall performance~\cite{duh2008ranking}) and 

%\begin{table}[t]
%\resizebox{\columnwidth}{!}{%
%    \centering
%    \begin{tabular}{c|cccccc|cccccc}
%\toprule
%        & 1 & 2 & 3 & 4 & 5 & 6 & 1 & 2 & 3 & 4 & 5 & 6\\\midrule
%        1 & - & 14 & 14 & 22 & 28 & 36 & - & 32 & 12 & 12 & 28 & 22\\
%        2 & 42$^\dag$ & - & 24 & 20 & 46$^\dag$ & 44$^\dag$ & 26 & - & 14 & 24 & 52$^\dag$ & 24\\
%        3 & 52$^\dag$ & 40$^\dag$ & - & 16 & 48$^\dag$ & 46$^\dag$ & 26$^\dag$ & 50$^\dag$ & - & 8 & 40$^\dag$ & 34$^\dag$\\
%        4 & 52$^\dag$ & 32 & 10 & - & 42$^\dag$ & 48$^\dag$ & 32$^\dag$ & 38 & 6 & - & 48$^\dag$ & 38$^\dag$\\
%        5 & 32 & 16 & 16 & 16 & - & 26 & 18 & 22 & 16 & 14 & - & 28\\
%        6 & 44 & 22 & 28 & 28 & 44$^\dag$ & - & 46$^\dag$ & 30 & 16 & 14 & 48$^\dag$ & -\\
    %     TA-Transfo &116.3M & 15.45 &\textbf{1.86}\\
%    \bottomrule
%    \end{tabular}
%}
%    \caption{Results of the pairwise preference test: the proportion of responses generation by system A (column) is more favoured than B (row). $^\dag$ means the preference is significant with $p<.05$.}
%    \label{tab:pt2}
%\end{table}

\subsubsection{Ranking based Human Evaluation}

We use pairwise binary ranking (i.e., preference test)~\cite{vilar2007human}, which has been shown reliable for comparing the performance of multiple models. 
%\RED{[CL: can we empathise the robustness of this ranking method?]}. 
We randomly sample 100 dialogue context from the test set for both single-turn and multi-turn dialogues (i.e., 50 samples each type). We then generate a response with each tested model given a sampled context. Given two responses generated by two models, two raters (PhD students in computer science) were asked to decide \textit{which model is better in terms of empathetic responding} or \textit{there is no difference}. 

\begin{table}[t!]
\resizebox{\columnwidth}{!}{%
    \centering \normalsize
    \begin{tabular}{lccc}
    \toprule
        Model  & Empathy & Relevance & Fluency \\\midrule
        MoEL & 1.629$^{\dag\ddag}$ & 1.525$^{\dag\ddag}$ & 2.031$^{\dag\ddag}$ \\
        Transfo & 1.987$^\dag$   & 2.043$^\dag$      & 2.307~ \\  
        PRE & 1.940$^{\dag}$   & 2.043$^{\dag}$      & 2.270~ \\
        MTL & 1.850$^{\dag}$    & 1.917$^{\dag}$      & 2.313~ \\\midrule
        ADM (ours) & 2.020$^{\dag}$    & 2.012$^{\dag}$      & 2.330~ \\
        AD (ours) & 2.177$^\ddag$    & 2.233$^\ddag$      & 2.380~ \\
        AD+DE (ours) & 2.187$^\ddag$    & 2.237$^\ddag$      & 2.387~ \\
    \bottomrule
    \end{tabular}
    }
    \caption{LSR Results. $^\dag$ means AD+DE significantly outperforms the corresponding model with $p<.05$ and $^\ddag$ means the corresponding model outperforms Transfo significantly with $p<.05$ (paired $t$-test).}
    \label{tab:likert}
\end{table}

\begin{table*}[t!]
    \centering \small
    \begin{tabular}{lcccccc}
    \toprule
        & \multicolumn{3}{c}{Single-turn} & \multicolumn{3}{c}{Multi-turn} \\
        \cmidrule(lr){2-4}\cmidrule(lr){5-7}
        Model & Empathy & Relevance & Fluency & Empathy & Relevance & Fluency \\\midrule
        MoEL & 1.714 & 1.652 & 2.050 & 1.543 & 1.397 & 2.012 \\
        Transfo & 2.073 & 2.193 & 2.373 & 1.900 & 1.893 & 2.240 \\
        Prepend & 1.920 & 2.047 & 2.327 & 1.960 & 2.040 & 2.213 \\
        MTL & 1.920 & 2.027 & 2.353 & 1.780 & 1.807 & 2.273 \\
        MTL+AD+DE & 2.067 & 2.093 & 2.340 & 1.973 & 1.933 & 2.320 \\
        AD & 2.187 & 2.260 & 2.373 & 2.187 & 2.213 & 2.400 \\
        AD+DE & 2.180 & 2.240 & 2.367 & 2.173 & 2.227 & 2.393 \\
    \bottomrule
    \end{tabular}
    \caption{LSR results for single-turn and multi-turn dailogues respectively.}
    \label{tab:lsr_full}
\end{table*}

We report the results of this pairwise preference test in Table~\ref{tab:pt}, and the corresponding break down results of the single-turn and multi-turn dialogues in Table~\ref{tab:pt_full}. Take the number 48 corresponding to Transfo and MoEL in Table~\ref{tab:pt} as an example. It means that 48\% of the judges prefer Transfer over MoEL by considering both single-turn and multi-turn dialogues in the test set. Table~\ref{tab:pt_full} gives the break down results for single-turn (i.e., 46\% ) and multi-turn dialogue (i.e.,  50\%), respectively. By taking the average, we can derive 48\% as the overall result.
%\textcolor{red}{[\textbf{TODO}: ** add an example to explain what the numbers mean, e.g., the 48 between Transfo and MoEL, and HOW this 48 can be derived from Table 4**]}
Clearly, human evaluation (i.e., Table~\ref{tab:pt}) shows very different observations compared to the automatic evaluation.  
On the one hand, AD and AD+DE are clear winners this time, which significantly outperform all other models including the best performed baseline \textit{PRE}. 
It can also be observed that AD+DE slightly outperformed AD but the difference is insignificant. 
On the other hand, multi-task learning shows a negative effect on empathetic dialogue modelling, i.e., by comparing MTL with Transfo and by comparing ADM with AD+DE. We give more discussions regarding this phenomenon in the Rating experiment section.  

In addition, it can be observed that MoEL gives the worst performance compared to all other models by a large margin, but one might argue that the results are not directly comparable because the non pre-trained MoEL has less capacity than other pre-trained baseline models (e.g., the parameters of Transfo are 5 times as many as that of MoEL).
%\textcolor{red}{[CL: IMPORTANT: need to articulate what do you mean by smaller model and why would this affect the model's performance]}
%On the other hand, models except the MTL and the MoEL significantly defeat the \textit{transfo} and new significant differences appear in the winning of AD and AD+DE against the \textit{prepend} model this time. 
The inconsistency on the results of automatic evaluation and preference test somewhat resemble the observation of prior studies that automatic metrics show low validity for evaluating empathetic dialogue systems~\cite{Liu2016e}. 
To further investigate the underlying issue, we interview our raters as to which factor influences most on their decisions. 
It turns out that small errors in the responses that cannot be detected by the automatic  measures (e.g. BLEU or BOW) can have a great impact. 
For instance, wrong reference (e.g., responding \emph{``I'm happy for you.''} when the speaker is actually describing an experience of his/her sister) or wrong tense (e.g., responding \emph{``I hope you will be fine.''} when the speaker is describing an experience happened in the past).
%More example outputs of each model can be found in Appendix~A.

%Besides, there is still no difference identified between AD and AD+DE, which denies the effectiveness of DE in term of improving the general performance of a EDS.

\subsubsection{Rating based Human Evaluation}

Likert Scale Rating (LSR) and Magnitude Estimation (ME) are two popular rating based methods. 
It is reported that ME performs better for evaluating goal-oriented dialogue systems~\cite{santhanam2019towards} and language generation systems~\cite{novikova2018rankme} while LSR works better for measuring acceptability of text~\cite{langsford2018quantifying}. 
Considering the degree of empathetic is tied to the acceptability of the generated responses and that multi-item LSR is on a par with ME~\cite{van-der-lee-etal-2019-best}, we opt for LSR with three dimensions listed below.
Model responses (the same set used in the ranking study) were scored by same two raters. 
The rating score ranges from 0 to 3. 
\begin{itemize}
    \item Empathy: \emph{Does the listener understand the speaker's feelings, and responds appropriately?}
    \item Relevance: \emph{Is the content of the reply relevant to the topic mentioned by the speaker? Is it informative?}
    \item Fluency: \emph{Does the response look fluent?}
\end{itemize}

\begin{table*}[t!]
    \centering 
    \begin{tabular}{ll}
    \toprule
         & Generated Response \\ \midrule
        \textbf{Input} & S: I got a free pizza from Domino's but felt bad about how it \\&happened.\\ \midrule
        Transfo  & oh no ! what did you do ? \\
        PRE & oh no , what happened ?  \\
        MTL & i love domino 's pizza ! i love domino 's ! \\ \hdashline
        ADM & oh no , what happened ? \\
        AD & why did you feel bad about it ? \\
        AD+DE & why did you feel bad about it ? \\ \midrule
        \textbf{Input} & \makecell[l]{S: Do any of your kids play any sports? \\ L: No, I do not have kids. Yours? \\ S: My daughter skateboards. She went down the big ramp the other \\ day. It was so cool.} \\ \midrule
        Transfo  & that sounds like a lot of fun !  \\
        PRE & wow , that must have been a lot of fun .   \\
        MTL & that 's cool . i 've never been on a rollercoaster . i 've never been on \\&a rollercoaster . \\ \hdashline
        ADM & that is so cool ! i bet you are so proud of her !  \\
        AD & that 's awesome ! i bet you were so proud ! \\
        AD+DE & that 's awesome ! i bet you were so proud of her ! \\
    \bottomrule
    \end{tabular}
    \caption{Example outputs generated by the baselines and our models, where S stands for speaker and L stands for Listener.}
    \label{tab:example}
\end{table*}

The rating results in Table~\ref{tab:likert} and Table~\ref{tab:lsr_full} (break down results) show a similar tendency with the ranking experiment and give some additional insights.
%\footnote{More break down results for single- and multi-turn dialogues for either ranking or rating experiment can be found in Appendix~C.%}
%}. 
Regarding \textit{fluency}, how or whether a model incorporates emotion information seems to have no impact to the fluency of the generated responses. In terms of the other two aspects, we have the following findings: 
(1) similar to the ranking experiment, MoEL gives the worst performance, regardless the rating aspect. Comparing to \textit{transfo}, the PRE, MTL, and ADM models cannot improved either \textit{empathy} or \textit{relevance} of the generated responses, significantly. The remarkable well performance of the vanila \textit{transfo} model embodies that by fine-turning the model on \textsc{EmpDial}, this GPT based model is a decent baseline for understanding emotion and responding empathetically;   
%First of all, similar to ranking experiment, MoEL gives the worst performance, regardless the rating aspect.  
%We, therefore, focus on other models in our analysis. 
(2) in line with the ranking experiment, AD and AD+DE give the best performance. 
Although AD+DE performs slightly better than AD, the difference between them is not significant. 
Joining with other results, it seems that learning separate embeddings for the speaker and listener does bring some benefit but it is not as strong as expected. 
Nonetheless, we found that introducing DE can help analyse the behaviours of listeners and speakers in terms of how they utter with respected to the same emotion situation, which will be discussed in detail in \S\ref{sec:interpret};
%the speaker and listener are distinguished when modelling emotion. 
%By looking at the nearest neighbours (NNs) of a emotion in the embedding space, we see another advantage of introducing DE, i.e., it can help analyse the behaviours of listeners and speakers in terms of how they utter with respected to the same emotion situation. 
%For example, for the emotion ``\textit{proud}'', its NNs in the speaker embedding space are words like \emph{honer}, \emph{happy}, and \emph{pleased} while the NNs in the listener space are words like \emph{celebrate}, \emph{keep}, and \emph{moment} (more  analysis of NNs is given in Appendix~D); 
(3) comparing the results of MTL with Transfo and that of ADM with AD+DE reveals that MTL decreases both empathy and relevance of the responses. % and probably should be excluded when designing an EDS. %\RED{[CL: can we add some explanations why MTL is not working here?]} 
%On the top of all these findings and findings of the ranking experiment, one additional insight is that besides showing empathy of speaker emotion, being relevance is also a key factor for empathetic responding, e.g., Prepend model has similar empathy score as well as higher relevance comparing to \emph{transfo}, but significantly defeat \textit{transfo} in preference test.
%\textcolor{blue}{
One possible reason behind why MTL does not yield positive effect in EDS (based on the results of both ranking and rating experiments) is that there might exist trade-off between the optimisation of the dialogue generator's objective and that of the emotion classifier's objective~\cite{sener2018multi}. As a result, the overall performance is harmed by the naive  linear combination of the two objectives.

\section{Case Studies}

\subsection{Sample Outputs of Different Empathetic Dialogue Systems}
\begin{table*}[t!]
     \centering
     \begin{tabular}{lp{10cm}}
     \toprule
     Emotion State &   Nearest neighbour words of the emotion label \\ \midrule
     \multirow{2}{*}{\textbf{Proud}} &  \textbf{S}: son, graduated, proud, honour, daughter, happy, pleased, nephew, musicians, said \\
     &  \textbf{L}: celebrate, bet, con, proud, keep, parent, started, moment, congratulations \\  \midrule
     \multirow{2}{*}{\textbf{Sad}} &  \textbf{S}: sad, cried, upset, bummed, died, passed, cry, depressed \\
     &  \textbf{L}: sorry, retrace, memories, sleazy, lose, toll, alive, sudden\\
     \bottomrule 
     \end{tabular}
     \caption{The 10 nearest neighbour words of the emotion label \textbf{PROUD} and \textbf{SAD} in the speaker (S) and listener (L) space, respectively.}
     \label{tab:visual}
 \end{table*}

Table~\ref{tab:example} lists a number of sample responses generated by the baselines and our models. It can be observed that our AD and AD+DE models produce high quality empathetic responses. For the first example, our models can follow the context and ask the reason why the speaker felt bad about getting a free pizza, whereas some of the baseline models produce uninformative responses (e.g., \emph{what did you do?}) and some of them respond with incorrect emotion (e.g., \emph{I love domino's pizza!}). In the second sample, our models can generate  more empathetic responses (i.e., providing more approval and praise) compared to other baselines.  In contrast, method like MTL generate irrelevant content (i.e., \emph{rollercoaster}). Another observation is that the responses generated by AD and AD+DE are quite similar to each other, which is in line with the evaluation results.

\subsection{Interpreting Dual Emotional Embeddings} \label{sec:interpret}

%\begin{figure}[t!]
%%    \centering
%    \includegraphics[width=\textwidth]{img/embed.jpg}
%    \caption{The 19 nearest neighbour words of the emotion label \textbf{PROUD} in the speaker (left column) and listener (right column) space respectively.}
%    \label{fig:visual}
%\end{figure}

% \begin{figure}[tb!]
% \begin{tcolorbox}[colback = white, boxrule = 0.3mm]
% [confident]\\
% S: tomorrow is going to be a great day\\
% L: what 's going to be a great day ?\\
% S: i have a job interview . i am sure i will get the job .\\
% L: that 's great ! i hope you get the job !
% \end{tcolorbox}
% \begin{tcolorbox}[colback = white, boxrule = 0.3mm]
% [afraid]\\
% S: tomorrow i am going to a haunted house .\\
% L: are you scared of haunted houses ?\\
% S: i am scared of them . i am scared of the dark .\\
% L: do you have a night light ?
% \end{tcolorbox}
% \caption{Given the initial word \emph{tomorrow}, two example dialogues are generated conditioning on the emotion labels ``confident'' and ``afraid''.}
% \label{fig:example_1}
% \end{figure}

%\textcolor{red}{[CL: *** need to do a much better job in this part, as it's a selling point for our model. Show the figures instead, which is more interesting.]}
We also conducted an experiment to assess how the learnt emotion embeddings by AD+DE differ with respect to speakers and listeners. Given an emotion label, we listed the label's top-10 nearest neighbours in the speaker space and listener space (see Table~\ref{tab:visual}), respectively, based on the label embedding. Take the emotion label ``proud'' as an example, words like \emph{proud}, \emph{happy}, \emph{honour} in the speaker space are very close semantically and are highly relevant to the emotion label. Also words like \emph{son}, \emph{daughter} are often be mentioned in parents' expression of pride. In the listener space, words like \emph{congratulations}, \emph{proud}, \emph{celebrate} are commonly used for responding to the speaker's emotion of proud and the corresponding experience. These examples not only show consistency with people's conversation habits, but also illustrate the difference between the speaker's and the listener's diction.

\subsection{Generating Empathetic Dialogues from Scratch}

Since we jointly model the speakers and listeners in the empathetic dialogues, our system is capable to generate a multi-turn conversation given an emotional situation and a prompt. Figure~\ref{fig:example_2} provides some example dialogues generated by AD+DE in such a way. 
After given a specific emotion label (e.g., \textit{joyful} and \textit{disappointed} from the predefined label set of the \textsc{EmpDial} dataset), our model can generate relevant and empathetic responses conditioned on the initial prompt such as ``my mother''. It can be observed that the generated multi-turn conversations are coherent and respect the given emotion labels. 
%In addition, our model also asks [???] follow-up topics to continue the conversation. 
%This shows that our model can capture the contextual history of the conversation in addition to respecting the given emotion label during the response generation process.}
\begin{figure}[tb]
\begin{tcolorbox}[colback = white, boxrule = 0.3mm, text width=0.85\columnwidth]
\textbf{[joyful]}\\
S: my mother just got a promotion at her job !\\
L: that 's great ! what kind of job is it ?\\
S: it 's a financial analyst job !\\
L: that 's great ! i 'm sure you 're proud of her !
\end{tcolorbox}
\begin{tcolorbox}[colback = white, boxrule = 0.3mm, text width=0.85\columnwidth]
\textbf{[disappointed]}\\
S: my mother was diagnosed with pan cre atic cancer a few weeks ago .\\
L: oh no ! i 'm sorry to hear that . is she going to be okay ?\\
S: i think so , but i was n't expecting it at all .\\
L: i 'm sorry to hear that . i hope everything works out for you .
\end{tcolorbox}
\caption{Given the initial word \emph{my mother}, two example dialogues are generated conditioning on the given emotion ``joyful'' and ``disappointed''.}
\label{fig:example_2}
\end{figure}

\section{Conclusion}

In this paper, we propose a simple and effective technique called Affective Decoding for empathetic response generation. 
%This method can additionally be augmented with an auxiliary dual emotion encoder, which learns separate embeddings for the speaker and listener given the emotion base of the dialogue.
%We assess different methods for emotion modelling in EDS by conducting both automatic evaluation and human evaluations.
Empirical results based on extensive human evaluation show that our models (AD and AD+DE) outperform several strong baselines. Simply fine-tune the pre-trained \textit{Transfo} on \textsc{EmpDial} achieves decent performance. MTL, which has been used in some EDS, shows negative effects on the overall performance. 
As a side outcome, we also confirm the low validity of the mainstream automatic metrics for evaluating empathetic dialogue systems. 

It was noted that empathetic dialogue systems tend to generate generic responses such as ``\textit{I'm sorry to hear that.}''. Therefore, one important future work is to improve the diversity and informativeness of the empathetic responses generated by an EDS. One possible technical direction is to employ variational autoenoders~\cite{zhao-etal-2017-learning,li-etal-2019-stable,li-etal-2020-improving-variational}, which have been shown effective in improving the diversity in response generation.  

\section*{Acknowledgement}
This work is supported by the awards made by the UK Engineering and Physical Sciences Research Council (EP/P011829/1) and Ningbo Natural Science Foundation (202003N4320, 202003N4321). 
We thank anonymous reviewers for their insightful comments. 

\bibliographystyle{acl_natbib}
\bibliography{anthology,acl2021}

%\appendix

\end{document}